\documentclass[conference]{IEEEtran}
\usepackage{caption}

\usepackage{cite}
\usepackage{times}
\usepackage{epsfig}
\usepackage{graphicx}
\usepackage{amsmath}
\usepackage{amssymb}
\usepackage{subcaption}
\usepackage[hyphens]{url}
\usepackage{hyperref}
\usepackage{tabularx}
\usepackage{multirow}
\usepackage{float}
\usepackage{adjustbox}

\ifCLASSINFOpdf

\else

\fi

\begin{document}

\title{End-to-End 3D Object Detection using LiDAR Point Cloud}

\author{\IEEEauthorblockN{Gaurav Raut}
\IEEEauthorblockA{University of Maryland\\
Email: gauraut14@gmail.com}
\and
\IEEEauthorblockN{Advait Patole}
\IEEEauthorblockA{University of Maryland\\
Email: advaitpatole@gmail.com}
}

\maketitle

\begin{abstract}
There has been significant progress made in the field of autonomous vehicles. Object detection and tracking are the primary tasks for any autonomous vehicle. The task of object detection in autonomous vehicles relies on a variety of sensors like cameras, and LiDAR. Although image features are typically preferred, numerous approaches take spatial data as input. Exploiting this information we present an approach wherein, using a novel encoding of the LiDAR point cloud we infer the location of different classes near the autonomous vehicles.
This approach does not implement a bird's eye view approach, which is generally applied for this application and thus saves the extensive pre-processing required. After studying the numerous networks and approaches used to solve this approach, we have implemented a novel model with the intention to inculcate their advantages and avoid their shortcomings. 
The output is predictions about the location and orientation of objects in the scene in form of 3D bounding boxes and labels of scene objects.
\end{abstract}

\IEEEpeerreviewmaketitle

\section{Introduction}
For autonomous vehicles, an important task for navigation is being aware of the objects in their surroundings. The environment for such a system is highly dynamic and surrounding objects can include pedestrians, other vehicles, or other stationary objects.
Over the past decades, extensive research has been done for the detection of objects in 2D. But with advancements in processing power, compute and sophisticated control systems, detecting 3D bounding boxes of surrounding objects have gained popularity.
To accurately record the sizes, orientations, and locations of objects in the real world, 3D object detection becomes necessary. Robotics, Augmented Reality (AR), self-driving automobiles, and other real-world applications could all benefit from the use of 3D detection with this new ability to perceive the world similarly to humans.
3D object detection enables autonomous vehicles to detect objects of interest with their 3D attributes and enables them to take actions as required. Most autonomous vehicles are equipped with 3D LiDAR and we plan on using this data to provide 3D object detection in our paper.
The task is challenging because of the large number of possible objects that may be present in a scene, and the varying poses, sizes, and appearances of these objects\cite{survey3dother}. In order to detect and classify these objects accurately, object detection algorithms must be able to handle a wide variety of data and scenarios. This makes it challenging to develop and evaluate object detection algorithms and requires using large, high-quality datasets like KITTI. \cite{3Dsurvey} \\
The KITTI 3D Object Detection dataset is a collection of data for 3D object detection in autonomous driving scenarios. It was developed by the Karlsruhe Institute of Technology and Toyota Technological Institute, and it is one of the most popular datasets for this task. The dataset includes images taken from a car-mounted camera, along with annotated 3D bounding boxes for various objects, such as cars, pedestrians, and road signs. The data is split into training and test sets, and it is commonly used to evaluate the performance of 3D object detection algorithms.

\subsection{Related Work}
Traditionally, object detection was done using hand-crafted features \cite{HOG} from image data, but with the onset of the CNN era, research has shifted towards feature learning\cite{wang2020overview}. Object detection was solved \cite{RoarNet} in two stages first, candidate proposals using the region proposal network (RPN) second being classification of these regions using a network.\\
The related work can also be divided into object detection using RGB images \cite{SingleShot}, using point cloud data \cite{EndtoEnd} and combined RGB images with point cloud \cite{Frustrum}\\
Here we explore the three state-of-the-art models which give exceptional accuracies and have inspired our novel approach.\\
\begin{enumerate}
    \item{\textbf{VoxelNet [2017]}} \cite{voxelnet}: The fundamental of VoxelNet is the idea of voxels, which are the three-dimensional analogues of pixels in a two-dimensional image. VoxelNet is a neural network model that uses 3D convolutional layers to learn spatial properties from data. It has been shown to be better than other approaches to detecting objects in 3D space. However, it can be computationally expensive and may not work well with objects that have complex shapes. It may also struggle to identify partially visible or obscured objects. Overall, VoxelNet is a promising approach for 3D object detection but has some limitations.
    \item{ \textbf{BirdNet+ [2018]}} : \cite{BirdNet,BirdNet+} The BirdNet+ model is an expanded version of the initial BirdNet model, which was designed for 2D object identification and segmentation. BirdNet+ is able to predict the location, orientation, and semantic class of objects in 3D space using a combination of 2D and 3D convolutional layers. It has been demonstrated to outperform previous approaches on benchmark datasets and has the potential for real-world applications. However, it may require a large amount of training data and be computationally expensive. It also faces difficulty identifying and segmenting partially visible or complex objects and does not perform well when applied to new contexts or object classes. There is a need for fine tuning for specific datasets or tasks.

    \item{ \textbf{PointPillars [2019]}}: \cite{pointpillars} Point Pillars is a method for representing 3D point cloud data that uses a unique encoding to divide the point cloud into vertical columns called point pillars. This encoding allows for the use of 2D convolutional neural networks, which makes the model more effective and scalable than approaches that use 3D convolutional networks. The point pillars encoding allows the model to learn spatial relationships between points and more accurately predict the position and orientation of objects. However, the encoding discards the original 3D structure of the point cloud, which may make it difficult for the model to recognize objects with intricate details or complicated shapes. Additionally, the vertical columns in the encoding may overlap and cause the model to lose spatial information, making it less useful for scenes with high point densities.
\end{enumerate}

\subsection{Dataset}
We are training our model on the KITTI 3D Object Detection benchmark. We make use of the Velodyne LiDAR point clouds, the left RGB images, labels and calib data to perform detection and provide inference in our pipeline. The KITTI 3D Object Detection dataset consists of 7481 training samples and 7518 testing samples. The calib file provides extrinsic parameters for the relation between LiDAR and the Camera frame and also contains the intrinsic parameters for both the LiDAR and Camera \cite{Geiger2012CVPR}. Finally, the labels file contains information on annotations and class labels for each sample.

\subsection{Exploratory Data Analysis}
From the \href{https://www.cvlibs.net/datasets/kitti/eval_object.php?obj_benchmark=3d}{KITTI 3D object detection dataset}, we are using the left camera RGB images, Camera Calibration dataset, Velodyne point clouds data and the training labels datasets. For The label files, the data is present in the following delimiter-separated format:\\
\\
\textbf{Label}: Describes the detected object class. It can take the following values (Car', 'Don't Care', 'Pedestrian', 'Van', 'Cyclist', 'Truck', 'Misc', 'Tram', 'Person Sitting')\\
    \textbf{Truncated}: Is a float value between 0-1 which indicated what ratio of the detected objects is within the frame.\\
     \textbf{Occluded}: Is an int value from 0 to 3 indicating fully visible, partly occluded, largely occluded and unknown respectively.\\
    \textbf{Alpha}: Angle of observation\\
    \textbf{Bounding Box coordinates}: Left, top, right and bottom coordinates of the object 2D bounding box.\\
     \textbf{3D dimensions}: Indicates the height, width and length of the objects.\\
   \textbf{Position}: Indicates the x,y,z coordinates of object in camera coordinates\\
   \textbf{Y-Rotation}: The ry value in camera coordinates.
 \\

    \begin{figure}[h]
	\includegraphics[width=0.5\textwidth]{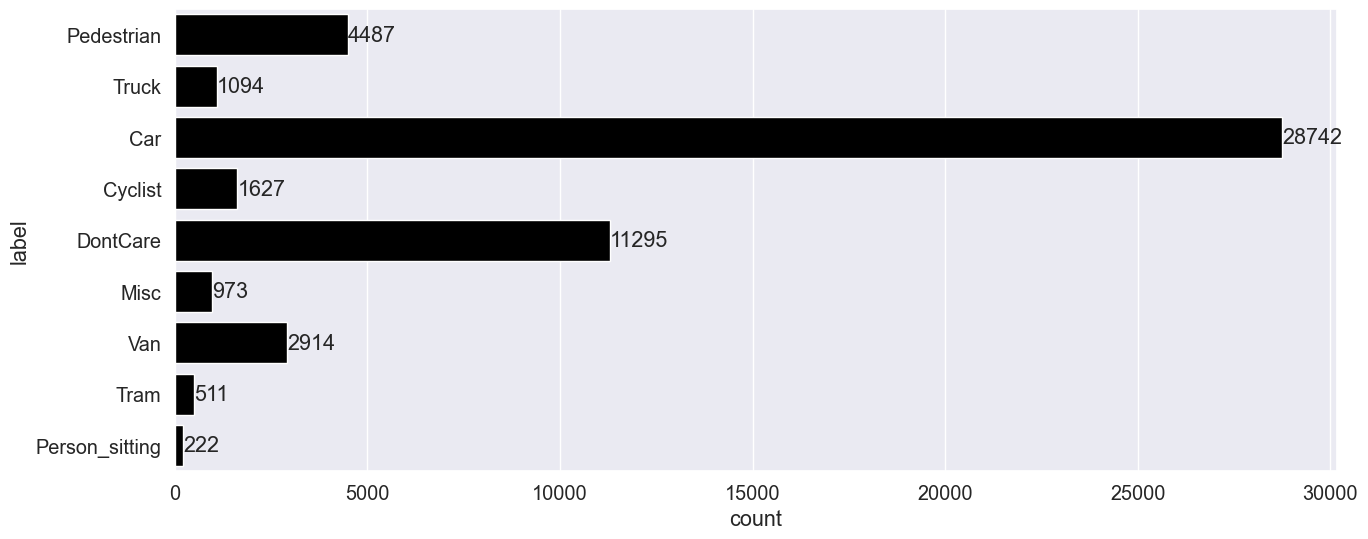}
	\caption{Training data category-wise count}
	\label{catcount}
\end{figure}

\section{Methods}
\subsection{Overview}
\begin{figure}[h]
	\includegraphics[width=0.5\textwidth]{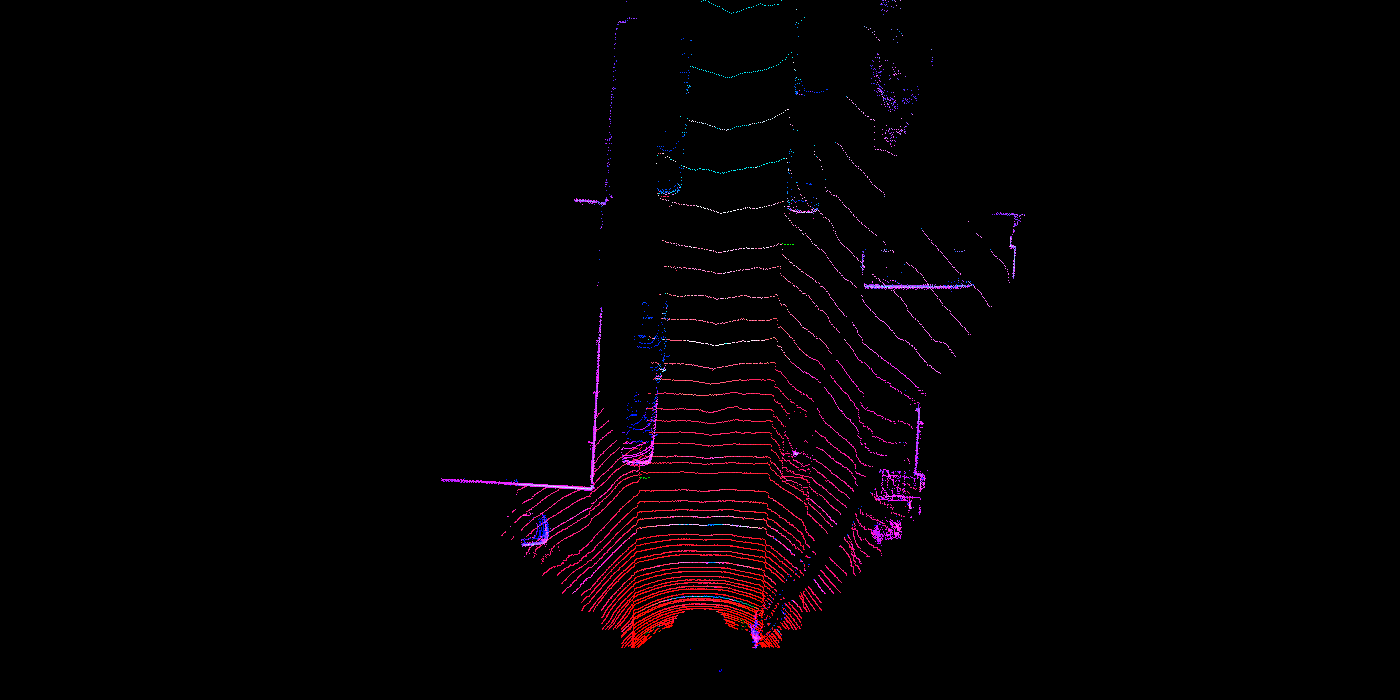}
	\caption{Generated Bird's Eye View image}
	\label{bev}
\end{figure}

The task of 3D object detection has many state-of-the-art implementations. We studied these implementations and devised two frameworks of our own. \cite{BirdNet+} made use of handcrafted Bird's Eye View images of point clouds which encoded height information of points in different colours. An example of the generated BEV is shown in figure \ref{bev}. Usually, pipelines using BEV images take this image, train an image detection network by adding 3D regression layers and predict the class, bounding box dimensions and the orientation angle. They share advantages from such image detection networks of being faster in inference speeds. Although, a major disadvantage is that such models are unable to properly detect the height of the objects since converting point clouds into a BEV image comes with the loss of structural information. Hence, we explored point cloud encoders for extracting point cloud features. We experimented with two state-of-the-art point cloud encoders which are the pillar feature encoder (PFE) \cite{pointpillars} and the voxel feature encoder (VFE) \cite{voxelnet}. Using both approaches, we constructed two powerful 3D detection models which achieve state-of-the-art performance on the KITTI 3D Object detection dataset.
\subsection{Method 1: PFE3DNet}
\subsubsection{Architecture}
\begin{figure*}[t]
	\includegraphics[width=\textwidth]{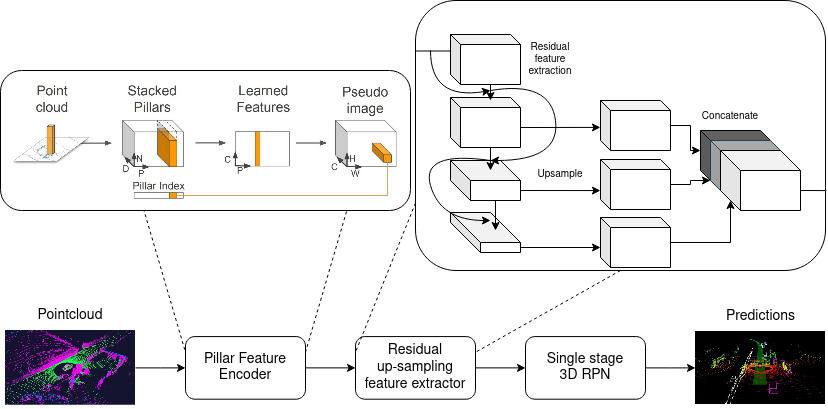}
	\caption{PFE3DNet Architecture}
	\label{pfe}
\end{figure*}

PFE3DNet uses the PFE layer for extracting features from the point clouds. The architecture is shown in figure \ref{pfe}. In this novel approach, point clouds are first voxelized in a 2D grid on the XY axes and then segregated as per their position on the grid. Later, each point is augmented by subtracting the mean of each grid cell ($x_c, y_c, z_c$) from itself. Finally, the mean of the 2D grid ($x_g, y_g$) is calculated and all of these features form a 9-dimensional ($D$) feature space for each point in the point cloud ($x, y, z, r, x-x_c, y-y_c, z-z_c, x_g, y_g$, where r is the reflectance). Finally, to convert it into a structured input for the network, the sparsity of the point cloud is exploited by capping the maximum number of non-empty pillars ($P$) and the maximum number of points per pillar($N$). Together, they form a pseudo image of $D\times P\times N$ dimensions and this image is passed to the further layers.\\

We have constructed a custom Residual Upsampling layer to extract features from the generated pseudo image. We have four blocks of convolutional layers which extract deeper features in a residual manner and are 49 layers deep. After the second block, features are decreased in their spatial resolution by a factor of 2 for every consecutive block transfer. The features from the second, third and fourth blocks are then upsampled and concatenated to be passed to the single-stage 3D RPN layer.\\
Finally, we make use of the 3D single-stage RPN layer to predict the dimensions, class and orientation of the 3D bounding boxes. We use the focal loss function to address the class imbalance issue.
\subsubsection{Weights and Loss Function}
We randomly initialize weights for all layers using kaiming initialization \cite{kaiming}.We have used the loss functions as they were introduced in \cite{voxelnet} \cite{pointpillars} as these are the standard loss functions used for the task of 3D object detection\cite{IOULoss}. Firstly, we will define the localization residuals which are defined as:
\begin{equation*}
	\begin{split}
		\Delta x = \frac{x_{gt} - x_{a}}{d_{a}}, \Delta y = \frac{y_{gt} - y_{a}}{d_{a}}, \Delta z = \frac{z_{gt} - z_{a}}{d_{a}} \\
		\Delta w = \log \frac{w_{gt}}{w_{a}}, \Delta l = \log \frac{l_{gt}}{l_{a}}, \Delta h = \log \frac{h_{gt}}{h_{a}} \\
		\Delta \theta = \sin{\theta_{gt}-\theta_{a}},
	\end{split}
\end{equation*}
where $gt$ and $a$ denotes the respective ground truth and anchor boxes and $d_{a} = \sqrt{w_{a}^{2}+l_{a}^2}$. All these terms add up to make the localization loss which is given as,
\begin{equation*}
	\mathcal{L}_{loc} = \sum_{b\in(x,y,z,w,l,h,\theta)} \mathrm{SmoothL1}(\Delta b)
\end{equation*}
A different loss function is used to distinguish the heading angle which is given as $\mathcal{L}_dir$. And finally, we use the focal loss to address the class imbalance issue as the classification loss. It is given as,
\begin{equation*}
	\mathcal{L}_{cls} = -\alpha_{a}(1-p_{a})^{\gamma}\log p_{a}
\end{equation*}
where $p_{a}$ is the anchor probability and $\alpha$=0.25, $\gamma$=2 as is mentioned in the above cited papers.
All of the above losses form our final loss function which is given as,
\begin{equation*}
	\mathcal{L} = \frac{1}{N}(\beta_{loc}\mathcal{L}_{loc}+\beta_{cls}\mathcal{L}_{cls}+\beta_{dir}\mathcal{L}_{dir})
\end{equation*}
where $N_{pos}$ are the number of positive anchors, $\beta_{loc}$=0.2, $\beta_{cls}=1$ and $\beta_{dir}=0.2$.
\subsubsection{Training}
The KITTI 3D Object Detection dataset consists of 7481 training samples and 7518 testing samples. The training data was split in a 50:50 ratio to make up for the validation set. We train the network for 60 epochs with a batch size of 1. The training was done on an i7 2.3 GHz CPU and RTX 3060 GPU. The validation was performed for every second epoch. A total of 250K iterations were performed on the training data. 
\begin{figure}[!h]
	\includegraphics[width=.5\textwidth]{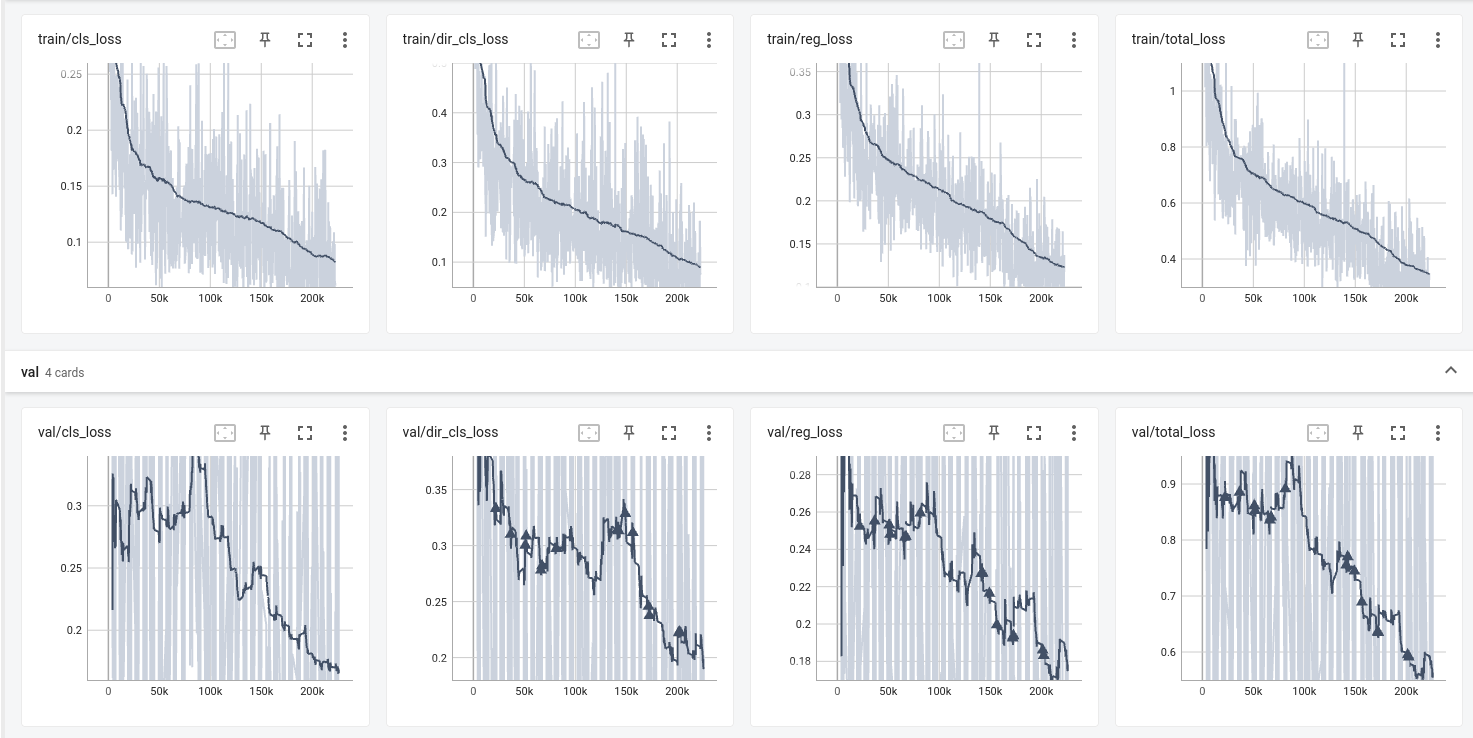}\vfill
	\includegraphics[width=.23\textwidth]{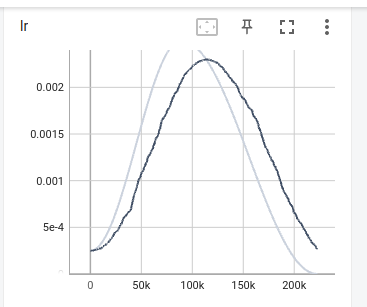} \hfill
	\includegraphics[width=.23\textwidth]{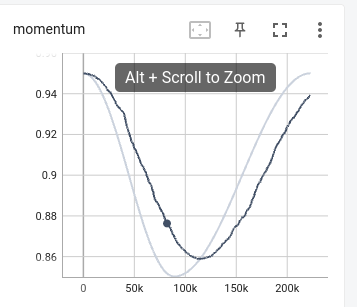}
	\caption{Training Graphs for PFE3DNet}
	\label{loss}
\end{figure}
\begin{figure}[h]
	\includegraphics[width=.5\textwidth]{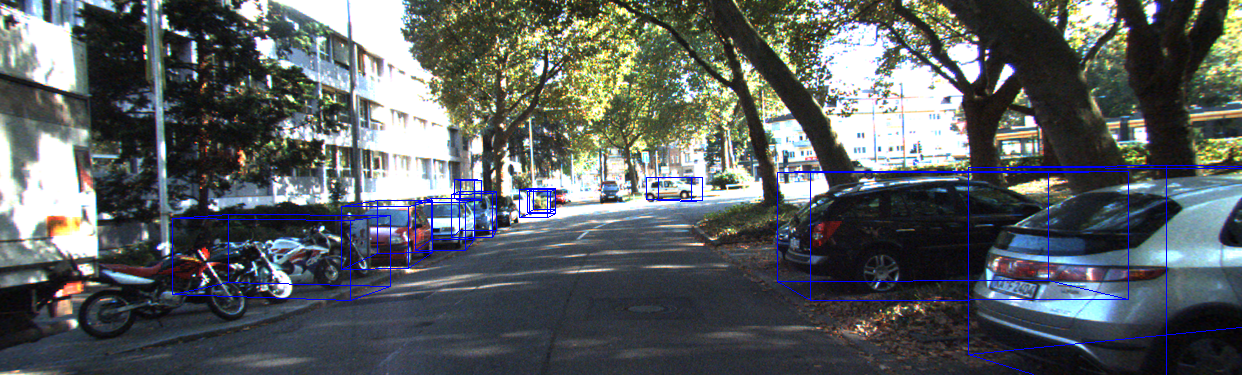}\vfill
    \includegraphics[width=.5\textwidth]{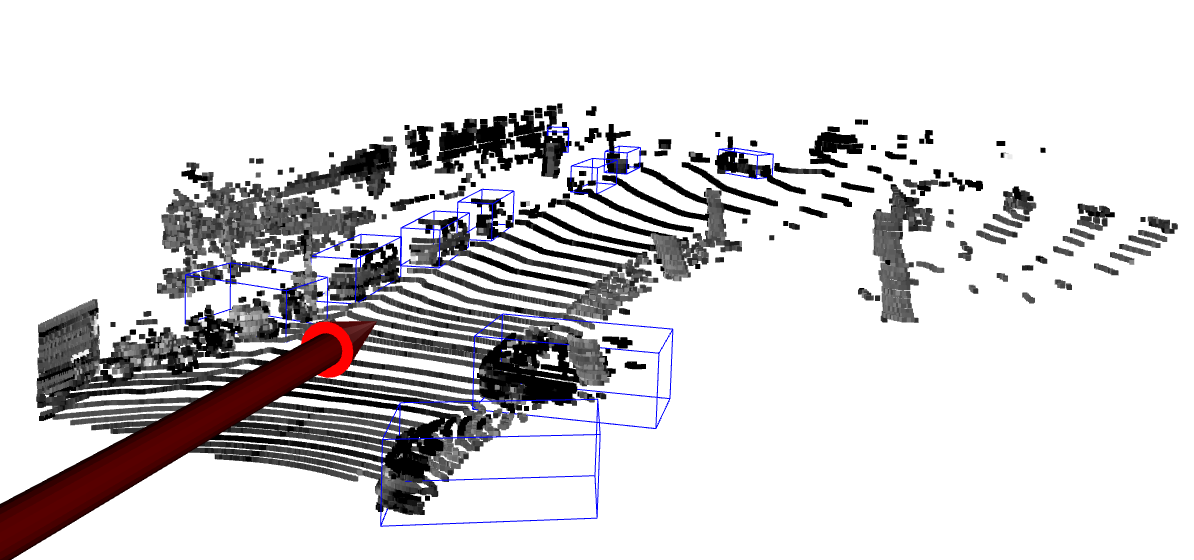}
	\caption{Output of the PFE3DNet}
	\label{op_pfe}
\end{figure}

\subsection{Method 2: Multi-Modal VFENet}
\subsubsection{Architecture}
\begin{figure}[t]
	\includegraphics[width=.5\textwidth]{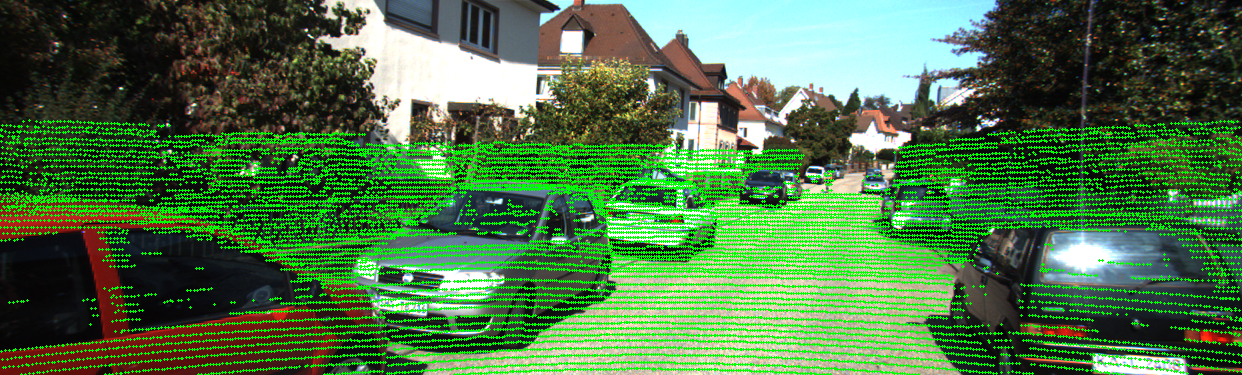}
	\caption{Fused point cloud and image data}
	\label{projected}
\end{figure}

Multi-Modal VFENet uses a multi-modal fusion of point cloud and RGB images. The architecture is shown in the figure. Here we have tried to extract image features from a convolution layer. In order to extract the image features we have tried the ResNet-50 backbone of an object detection network which was pre-trained on ImageNet. In order to extract features from the point cloud we tried to voxelize 
the point cloud so that we could get structured data in the form of grids. We divided the 3D point cloud into equally spaced voxel that covers the entire point cloud with range $D$, $H$ and $W$.We have selected the size of voxel as $v_{D}$, $v_{H}$ and $v_{W}$ hence the size of of voxel grid is $D'= D/v_{D}$, $H'= H/v_{H}$ and $W'= W/v_{W}$.In another half of the network, we try to extract the image features combined with the point cloud data. In order to fuse the point cloud data and image data we have made use of calibration matrices of the camera and the LiDAR \ref{projected}.

As discussed earlier we have made use of the backbone ResNet-50 architecture of pre-trained 2D detection model (Faster-RCNN)\cite{faster}, we passed this combined image to the ResNet-50 backbone network, but we have not completely used the entire network but have just made use of 2D convolution layers and have used the output feature maps of the conv4 layer. The output dimension of that layer is 512.We have reduced the dimension from 512 to 96 and then to 16 dimensions using fully connected layers. The image feature map is then concatenated with the point-wise features, which are fed to voxel feature encoding layers. In order to use the fused data as features we tried to create encodings for each voxel in the voxel. We created a voxel feature encoding, for example, if there is a voxel V = \{$p_{i}$ = ($x_{i}$, $y_{i}$, $z_{i}$, $r_{i}$) $\in$ $\mathrm{R}^{4}$ \}.We got the mean for each voxel V denoted as \{$v_{x}$, $v_{y}$, $v_{z}$\}.Then  we apply offset to each point $p_{i}$, hence the feature set becomes $V_{in}$ = ($x_i, y_i, z_i, r_i, x-x_i, y-y_i, z-z_i$), the point features are passed to the FCN which is composed of a linear layer, a batch normalization layer, and a rectified linear unit layer. After obtaining point-wise feature representations, we use element-wise MaxPooling across all point features associated to V to get the locally aggregated feature.These layers are named as VFE layers which is a concatenation of point features  and image features which are of 7 dimension and there are 16 dimensions from the CNN features whose dimension is reduced from 512 to 16.We use VFE-i($cin$, $cout$) to represent the i-th VFE layer that transforms input features of dimension $cin$ into output features of dimension $cout$. The linear layer learns a matrix of size $cin$ ×($cout$/2), and the point-wise concatenation
yields the output of dimension $cout$.Because the output feature combines both point-wise
features and locally aggregated feature, stacking VFE layers encodes point interactions within a voxel and enables the final feature representation to learn descriptive shape information.The output voxel features is represented as sparse tensor of dimension $C$ x $D'$ x $H'$ x $W'$ where $C$ is the dimension of the voxel wise feature.In the next stage we have used convolutional middle layers to aggregated voxel wise features using 3D convolution, batchnorm and ReLU layer.In the last stage we have made use of a region proposal network to get the region proposals output, the 
input to this network is from the convolution middle layers, and finally from the feature map we get two types of outputs: 1.probability score map 2. regression map.

\begin{figure*}[t]
	\includegraphics[width=\textwidth]{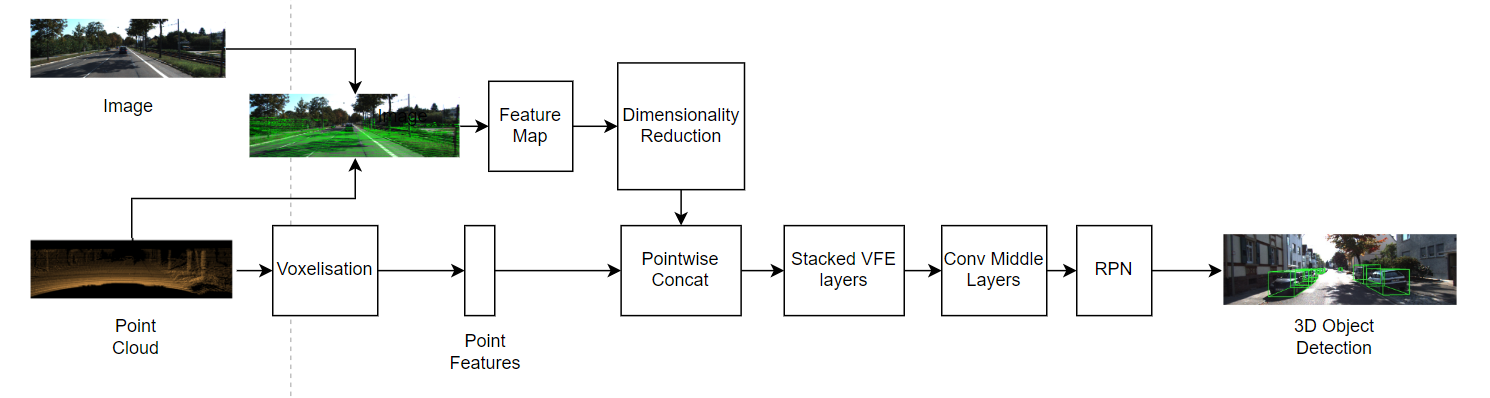}
	\caption{Architecture of Multi-Modal VFENet}
	\label{flowvfe}
\end{figure*}

\subsubsection{Loss Function} 
We have introduced our weights for the layers using Xavier initialization \cite{xavier}. We have adopted a similar loss function as mentioned in \cite{IOULoss}. Here we have parameterized our 3D bounding box with 7 parameters :($x_{c}, y_{c}, z_{c}, l_{c}, w_{c}, h_{c}, \theta_{c}$) which are the x, y, z centre coordinate and l, w, h are the length, height and width of the box and $\theta$ the yaw angle. To retrieve a positive anchor from ground truth we define the residual vector as  
\begin{equation*}
	\begin{split}
	&	\Delta x = \frac{x_{gt} - x_{a}}{d_{a}}, \Delta y = \frac{y_{gt} - y_{a}}{d_{a}}, \Delta z = \frac{z_{gt} - z_{a}}{d_{a}} \\
	&	\Delta w = \log \frac{w_{gt}}{w_{a}}, \Delta l = \log \frac{l_{gt}}{l_{a}}, \Delta h = \log \frac{h_{gt}}{h_{a}} \\
	&	\Delta \theta = {\theta_{gt}-\theta_{a}},
	\end{split}
\end{equation*}
where $gt$ and $a$ denotes the respective ground truth and anchor boxes and $d_{a} = \sqrt{w_{a}^{2}+l_{a}^2}$.The loss function is defined as : 

% \begin{equation*}{
% \begin{split}{
%  & {L = \alpha\frac{1}{N_{pos}}\sum_{i}^{}L_{cls}(p_{i}^{pos},1) +  \beta\frac{1}{N_{neg}}\sum_{i}^{}L_{cls}(p_{j}^{neg},0) + \\
%  & \frac{1}{N_{pos}}\sum_{i}^{}L_{reg}(u_{i}, u_i^{*})}	
% }\end{split}
% }
% \end{equation*}

\begin{equation*}
\begin{split}
& L = \alpha\frac{1}{N_{pos}}\sum_{i}^{}L_{cls}(p_{i}^{pos},1) +  \beta\frac{1}{N_{neg}}\sum_{i}^{}L_{cls}(p_{j}^{neg},0) + \\
& \frac{1}{N_{pos}}\sum_{i}^{}L_{reg}(u_{i}, u_i^{*})
\end{split}
\end{equation*}

where $p_{j}^{neg}$ and $p_{i}^{pos}$ are the softmax output of the negative and positive anchors, while  $u_{i}$ and $u_i^{*}$ are the regression output and the ground truth for the positive anchor. The $L_{cls}$ stands for the binary cross entropy loss and $L_{reg}$ is the regression loss in which we have used the Smooth L1 function \cite{Huber2011}.

\subsection{Training}
For training the network we have used the KITTI 3D object detection dataset. There are 7481 samples which are split as 3712 as the training set and 3769 samples as the validation, set. There are 3 different levels of difficulty namely: easy, moderate and difficult which are based on object size, occlusion and truncation. For training, we used Adam optimizer and then we added a scheduler to decay the learning rate, the initial learning rate is set as 0.003 and we try to reduce the learning by a factor of 0.1 for every 10 epochs. We trained our model with 80 epochs and a batch size of 2 and it took around 40 hours (30 minutes for each epoch) to train the model on a NVIDIA 3060 GPU  and an i7 2.3GHz CPU.As a result of which we could not experiment with it much as the model was very heavy because of 3D convolutions. We tried to train the network but to the 3D convolutions, we could not perform experiments on it.

\begin{figure}[!h]
	\includegraphics[width=0.23\textwidth]{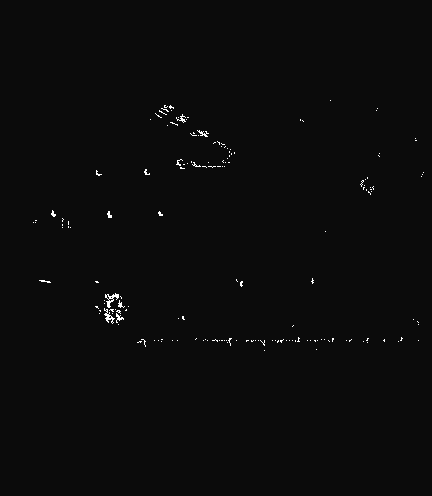}\hfill
	\includegraphics[width=0.23\textwidth]{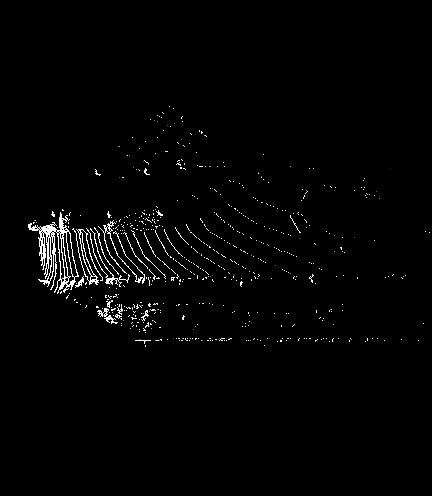}\vfill
	\includegraphics[width=0.23\textwidth]{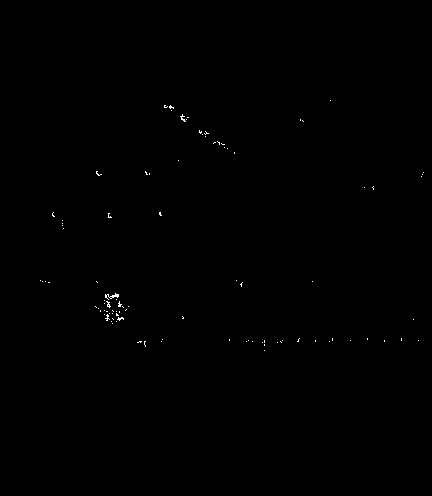}\hfill
	\includegraphics[width=0.23\textwidth]{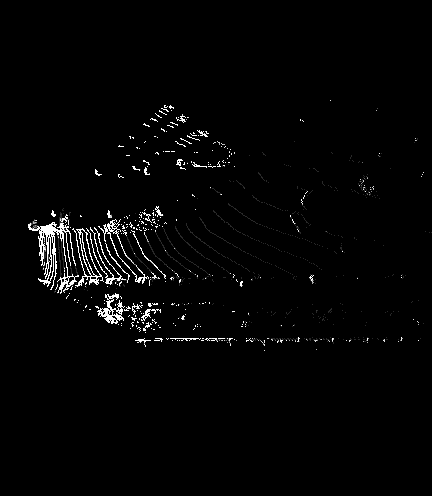}
	\caption{Visualization of pillar encoder features on different levels}
	\label{pfe_features}
\end{figure}
\begin{figure}[h]
	\includegraphics[width=.5\textwidth]{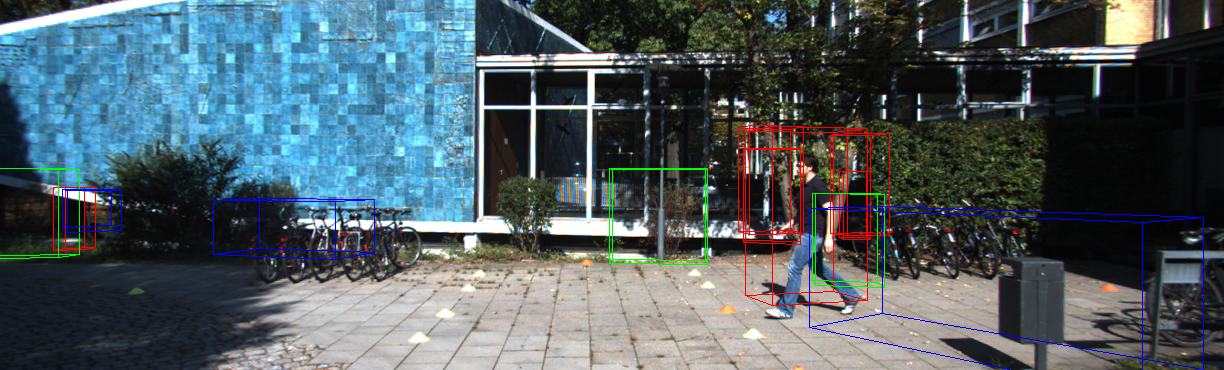}\vfill
    \includegraphics[width=.5\textwidth]{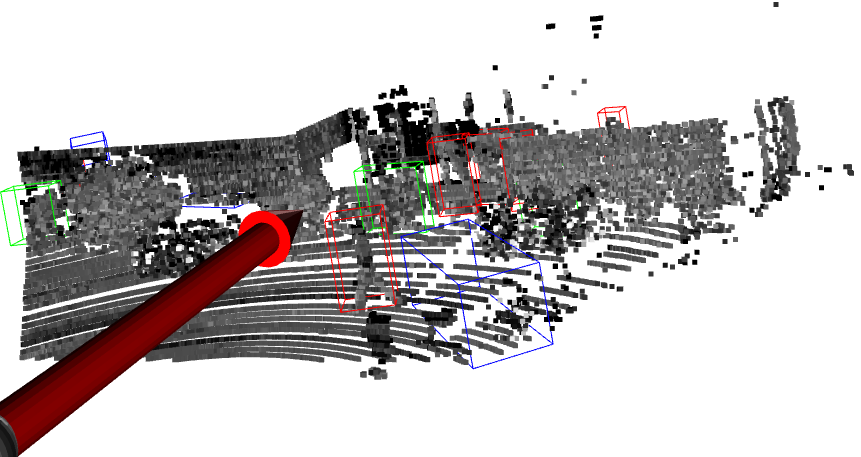}
	\caption{Output of the multi-modal VFENet}
	\label{op_vfe}
\end{figure}

\section{Experiments}
\subsection{PFE3DNet}
We started by implementing our initial pipeline of BirdNet+ and realized that even if the inference and training were faster, it wasn't the right way to deal with point clouds. Hence, we experimented with using point cloud encoders as opposed to BEV images. Examples from the pillar encoders are shown in the figure \ref{pfe_features}.
\begin{figure}[!h]
	\includegraphics[width=0.50\textwidth]{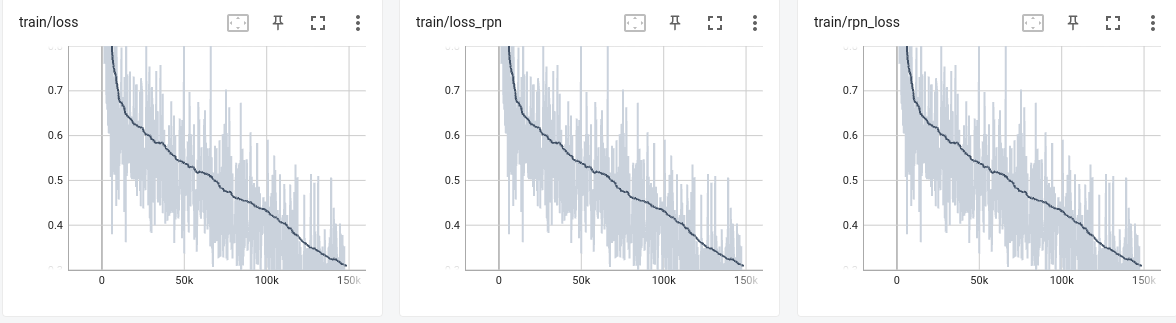}\vfill
	\includegraphics[width=0.50\textwidth]{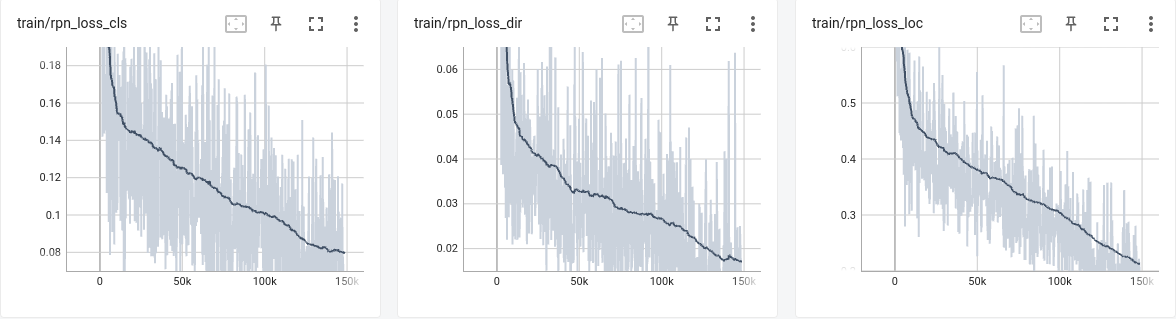}
	\caption{Training Graphs for Multi Modal VFE}
	\label{vfe_training}
\end{figure}
Since we are using Adam optimizer \cite{adam}, we started with our initial learning rate of $2.5\mathrm{e}^{-4}$ with a decay rate of $0.8\times$learning rate. With our initial training graphs, we saw that our learning rate jumps to about 0.002 and hence, in another training iteration, we started by setting the initial learning rate of 0.002 and a decay rate is reduced by 5 percent  after 10 epochs, hence building a faster momentum.
The loss increased and hence we reduced the learning rate for the final model.

\subsection{Evaluation metric}
We follow the KITTI 3D Object detection evaluation metrics to score our model against the baselines. We calculate the mAP (mean Average Precision) for different evaluation metrics such as the BBOX\_2D, BBOX\_BEV, BBOX\_3D and compare it with our baselines. Based on these evaluation metrics, we prove that the PFE3DNet and Multi-Modal VFE models outperform the baselines and achieve state-of-the-art 3D object detection results.

\begin{figure*}[t]
    \begin{tabularx}{1\textwidth} { 
  | >{\raggedright\arraybackslash}X 
  | >{\centering\arraybackslash}X 
  | >{\centering\arraybackslash}X
  | >{\centering\arraybackslash}X
  | >{\centering\arraybackslash}X
  | >{\centering\arraybackslash}X
  | >{\centering\arraybackslash}X
  | >{\centering\arraybackslash}X
  | >{\centering\arraybackslash}X
  | >{\centering\arraybackslash}X
  | >{\centering\arraybackslash}X
  | >{\centering\arraybackslash}X
  | >{\centering\arraybackslash}X
  | >{\raggedleft\arraybackslash}X | }
 \hline
\multirow{2}{*}{Method} & \multirow{2}{*}{Modality} & \multirow{2}{*}{mAP} & \multicolumn{3}{|c|}{Car} & \multicolumn{3}{|c|}{Pedestrian} & \multicolumn{3}{|c|}{Cyclist} \vline\\

 &  &  & Easy & Mod. & Hard & Easy & Mod. & Hard & Easy & Mod. & Hard\\
\hline
\small VoxelNet  & LiDAR & 49.05 & 77.47    & 65.11   & 57.73     & 39.48 & 33.69 & 31.5     & 61.22 & 48.36 & 44.37   \\
\hline
\small Point Pillars  & LiDAR & 59.20 & 79.05    & 74.99   & 68.30     & 52.08 & 43.53 & 41.49     & 75.78 & 59.07 & 52.92   \\
\hline
\small SECOND  & LiDAR & 56.69 & 83.13    & 73.66   & 66.20     & 51.07 & 42.56 & 37.29     & 70.51 & 53.85 & 46.90   \\
\hline
\small BirdNet+  & LiDAR (BEV) & 47.66 & 70.14    & 51.85   & 50.03     & 37.99 & 31.46 & 29.46     & 67.38 & 47.72 & 42.89   \\
\hline
\small Multi Modal VFENet  & LiDAR & 54.64 &83.2    &72.7   &65.2     & 40.18 & 34.88 &34.7     & 63.6 & 49.77 & 47.56   \\
\hline
\small PFE3D- Net (ours)  & LiDAR & \textbf{64.81} & \textbf{85.07}    & \textbf{76.77}   & \textbf{74.36}     & \textbf{52.91} & \textbf{47.47} & \textbf{44.50}     & \textbf{81.30} & \textbf{61.90} & \textbf{59.01}   \\
\hline
\end{tabularx}
\caption{Evaluation comparisons on 3D KITTI Test}
\label{table}
\end{figure*}

\subsection{Expected results}
Through this approach, we will be able to classify and locate different classes like cars, pedestrians, and cyclists and locate them. We will be validating our model by performing some experiments on the KITTI object detection benchmark focusing on 3D and Bird's Eye View detection tasks. The 3D and BEV detection results obtained by the proposed approach are comparable to the ones provided by the other state-of-the-art methods. However, the proposed framework is the only one among those presented here
that can estimate 3D boxes for all the evaluated categories using only LIDAR BEV images. Unlike other methods, our framework does not rely on additional sources of data (e.g. images) and is designed to perform multi-class detection using a single model and just one BEV representation (with fixed
grid resolution).

\section{Limitations}
The are limitations are as follows:
\begin{itemize}
	\item Both approaches rely heavily on tensor computations and hence require a GPU for faster inference.
	\item They are only trained on 3 different classes which is not sufficient as there are other obstacles present on roads as well.
	\item The VFE features create sparse tensors and try to perform 3D convolutions on them which is computationally heavy.
	\item Because of the sparse of the point clouds smaller points of interest with fewer points is difficult to be detected using point clouds.
\end{itemize}

\section{Conclusion}
We implemented two novel models using two different point cloud encoders. We also proved that our implementations perform than the baselines and are also able to achieve state-of-the-art performance. We also conclude that BEV images are an inefficient way to detect 3D bounding boxes since they lack spatial information as well as structural information about the point clouds. In the future, we plan on making point cloud encoding more efficient and we are also looking into the idea of using RGB images in order to extract useful features that can b more useful for the task of 3D object detection. We conclude that because of infusing image data in the VFE model, the model performed better than its baseline. We also conclude that our novel Residual Upsampling layer and the 3D Single Stage RPN in the PFE3DNet are better suited to extract features from the point cloud and are not only able to achieve better performance against the baseline but also achieve a SOTA output on the KITTI 3D detection test set.
{
	% \bibliographystyle{ieee}
	% \bibliography{egbib}
        \bibliographystyle{IEEEtran}
        \bibliography{End-to-End_3D_Object_Detection_using_LiDAR_Point_Cloud}
}

\end{document}